\documentclass[runningheads]{llncs}
\usepackage{url}

\usepackage{graphicx} % Required for inserting images
\usepackage{booktabs}       % professional-quality tables
\usepackage{xcolor}
\usepackage{float}
\usepackage{comment}
\usepackage{amsmath}
\usepackage{siunitx}
\usepackage[rightcaption]{sidecap}
\usepackage{subcaption}

\begin{document}

\title{Deep Learning–Assisted Detection of Sarcopenia in Cross-Sectional Computed Tomography Imaging}
\author{
Manish Bhardwaj\inst{1} \and
Huizhi Liang\inst{1} \and
Ashwin Sivaharan\inst{2}
Sandip Nandhra, \inst{2} \and
Vaclav Snasel\inst{3}\and
Tamer El-Sayed\inst{2} \and
Varun Ojha\inst{1}
}

\authorrunning{Bhardwaj et al.}

\titlerunning{Deep Learning–Assisted Detection of Sarcopenia}

\institute{Newcastle University, Newcastle, UK \and
The Newcastle upon Tyne Hospital NHS Foundation Trust, Newcastle, UK
\and
Technical University of Ostrava, Ostrava, Czech Republic
}
%\date{December 2024}

\maketitle

\begin{abstract}
Sarcopenia is a progressive loss of muscle mass and function linked to poor surgical outcomes such as prolonged hospital stays, impaired mobility, and increased mortality. Although it can be assessed through cross-sectional imaging by measuring skeletal muscle area (SMA), the process is time-consuming and adds to clinical workloads, limiting timely detection and management; however, this process could become more efficient and scalable with the assistance of artificial intelligence applications. This paper presents high-quality three-dimensional cross-sectional computed tomography (CT) images of patients with sarcopenia collected at the Freeman Hospital, Newcastle upon Tyne Hospitals NHS Foundation Trust. Expert clinicians manually annotated the SMA at the third lumbar vertebra, generating precise segmentation masks. We develop deep-learning models to measure SMA in CT images and automate this task. Our methodology employed transfer learning and self-supervised learning approaches using labelled and unlabeled CT scan datasets. While we developed qualitative assessment models for detecting sarcopenia, we observed that the quantitative assessment of SMA is more precise and informative. This approach also mitigates the issue of class imbalance and limited data availability. Our model predicted the SMA, on average, with an error of $\pm$3 percentage points against the manually measured SMA. The average dice similarity coefficient of the predicted masks was 93\%. Our results, therefore, show a pathway to full automation of sarcopenia assessment and detection.   

\keywords{Sarcopenia Assessment \and Skeletal Muscle Area \and Medical Imaging \and Computed Tomography Imaging \and Transfer Learning \and  Self-Supervised Learning}

\end{abstract}

\section{Introduction}
\label{sec:Introduction}
Sarcopenia is a progressive disorder that is associated with loss of muscle strength, muscle mass, and muscle function~\cite{cruz2019sarcopenia}. It becomes increasingly common as a person ages and is closely related to the physical frailty syndrome. Almost 30\% of the worldwide population over the age of 60 years is estimated to have sarcopenia~\cite{sarcoprevalence} and to exacerbate the scale of this problem, the ageing global population has led to the estimate that over 2 billion people will be over the age of 60 by 2025~\cite{sarcoagingpop}. Despite this, sarcopenia is considered an under-diagnosed condition, not least due to the variability in diagnostic criteria across multiple guidelines~\cite{cruz2019sarcopenia} and populations~\cite{AWGSOP2020}. Sarcopenia has been associated with worse survival in those undergoing treatment for multiple types of cancer, including hepatocellular carcinoma,~\cite{sarcoHCC} urothelial carcinoma, and gastric cancer~\cite{sarcoGastric}. Sarcopenia has also been associated with adverse surgical outcomes, including prolonged hospital stay, poor physical mobility and/or death~\cite{sarcoVascLL}. As a result, sarcopenia has become a focus of intense research aiming to improve diagnosis and treatment~\cite{santilli2014clinical}. Pre-operative sarcopenia assessment and optimization can significantly improve surgical outcomes. 

%%%Sarcopenia has also been associated with poor outcomes in surgical patients in multiple specialties; it has been shown that sarcopenic patients undergoing open vascular reconstruction for peripheral arterial disease (PAD), have poorer amputation-free survival~\cite{sarcoVascLL}. Meta-analysis on those with Sarcopenia, undergoing abdominal surgery have increased rates of complications and poorer overall survival~\cite{sarcoAbdo}. 

The most recent European Working Group on Sarcopenia in Older People (EWGSOP) guidelines suggest measuring muscle quantity or quality, using various cross-sectional imaging modalities, to confirm a diagnosis of sarcopenia~\cite{cruz2019sarcopenia}; an example of this is a measurement of skeletal muscle area (SMA) performed at the level of the third lumbar vertebra (L3) using computed tomography (CT) scans. This technique can be time-consuming; it requires identifying the correct slice in the transverse plane and manually measuring the visible skeletal muscle area (SMA) in that slice. As a result, despite many surgical patients undergoing CT scans, this valuable data source remains underutilised. Deep learning (DL) methodologies that measure SMA and identify sarcopenia can automate this process. 

However, DL has its challenges; a supervised deep-learning algorithm that can discriminate between sarcopenic and non-sarcopenic patients requires a large amount of labelled training data. In addition, data imbalance is a critical problem when dealing with supervised learning approaches. Thus, an isolated image classification implementation is insufficient to detect sarcopenia, as it would still require manual measurement. Therefore, we adopted the SMA assessment approach. The contribution of this paper is as follows:
\begin{itemize}
    \item[$\bullet$] We contribute an original high-quality dataset\footnote{Dataset: https://doi.org/10.5281/zenodo.15619502}. of computed tomography (CT) scans collected at NHS Freeman Hospital to detect sarcopenic and non-sarcopenic.
    
    \item[$\bullet$] We propose a self-supervised learning approach to estimate the skeletal muscle area at the third lumbar vertebra (L3) level from a CT scan. 

    \item[$\bullet$] We show that even with a limited dataset, our method predicted the skeletal muscle area in the range of 1\% to 7\% (on an average of 3\%) against the manually measured muscle area.
\end{itemize}

The rest of the paper is organised as follows. Sec.~\ref{sec:relatedwork} describes the related work followed by the proposed methodology in Sec.~\ref{sec:proposedapproach}. Experiment setting and results are discussed in Sec.~\ref{sec:results}. Sec.~\ref{sec:con} conclude this work.

\newpage
\section{Related work}
\label{sec:relatedwork}
{\color{black} Extensive studies have been carried out for the diagnosis of sarcopenia~\cite{turimov2023machine}. This also includes the application of machine learning algorithms using various health parameters. Because of its high prevalence, it is suggested to perform screening for older patients and those facing certain health conditions~\cite{fielding2011sarcopenia,liao2023use}. Sarcopenia has been referred to as an underdiagnosed condition, and the application of machine learning on CT scans provides the opportunity not to miss this condition~\cite{aali2024underdiagnosedmed}.} 

Transfer Learning (TL) and Self-supervised learning (SSL) approaches are the most effective methodologies used in the literature to mitigate the constraints of limited labelled datasets~\cite{zhao2024comparison}. TL aims to transfer the knowledge contained in a pre-trained network for different but related domains to the target learner~\cite{zhao2024comparison}, thus reducing the requirement of large data for training. For transfer learning, we identified the RadImageNet network, which has been pre-trained on multiple imaging modalities in multiple pathological conditions and anatomical locations~\cite{mei2022radimagenet}. Most works dealing with sarcopenia using deep learning use different imaging modalities (for example, x-ray ~\cite{ryu2023chest} or scans of different anatomical locations, such as the neck~\cite{ye2023development}. Gu et al.~\cite{gu2023detection} studied CT scans at the L3 vertebra using segmentation networks such as U-Net for predicting masks and comparing true and predicted SMA. Our approach follows a similar methodology but employs a self-supervised learning and Vision Transformer approach for predicting masks and calculating SMA. Moreover, no information on data processing is offered in Gu et al.~\cite {gu2023detection}, while our work follows European sarcopenia guidelines and clinicians' recommendations.~\cite{cruz2019sarcopenia}

Self-supervised learning reduces the requirement of a large amount of training data. We studied various SSL approaches belonging to SSL such as~\cite{he2020momentum},  \cite{punn2022bt}, \cite{jiang2022self} and \cite{dawid2023intro}. We observed that SMIT~\cite{jiang2022self} has a teacher-student network for learning various types of medical images, including CT scans. These images and scans were pre-processed in 2-dimensional and 3-dimensional space before being utilised for network training purposes.% \cite{shivdeo2021comparative}.

\section{Method} 
\label{sec:proposedapproach}
We propose a deep learning approach to automatically predict SMA at the level of L3 in CT images. To train such a deep learning network, a clinician must manually annotate and label a large number of images to outline the muscle area on the CT scan.  This can often be time-consuming and resource-intensive, and therefore, there is a lack of appropriate training data that can be used for DL methodologies. Given the limitations of these small datasets that come in two batches and two different formats, we explore two DL techniques. (i) A transfer learning approach, where an existing pre-trained model on a large set of medical images was fine-tuned using an inductive transfer learning methodology. We use RadImageNet~\cite{mei2022radimagenet} and fine-tune this with a small set of labelled datasets. (ii) a self-supervised learning (SSL) approach that alleviates the necessity of large labelled datasets. We employ self-distillation learning with masked image modelling to perform SSL for vision Transformers (SMIT)~\cite{jiang2022self}.  The detailed steps of our methodology are as follows.

\subsection{Sarcopenia assessment}
One of the radiologically validated methods for assessing sarcopenia involves measuring the SMA at the level of the L3 vertebra using CT scans. This anatomical landmark has been shown to correlate strongly with whole-body muscle mass and is widely used in both clinical and research settings for diagnosing sarcopenia~\cite{cruz2019sarcopenia}. However, this technique is time-consuming and labour-intensive,  posing a challenge for widespread implementation in busy healthcare settings. It requires identifying the correct CT slice in the transverse plane and manually measuring the area of the skeletal muscle visible in that slice. Therefore, despite many surgical patients undergoing CT scans, this valuable source data remains untapped for early detection of sarcopenia that could improve surgical outcomes. Automating the measurement of L3 SMA from CTs is an opportunity to improve patient outcomes through the timely identification and management of sarcopenia. 

\begin{figure}[h!]
        \centering
        \includegraphics[width=\textwidth]{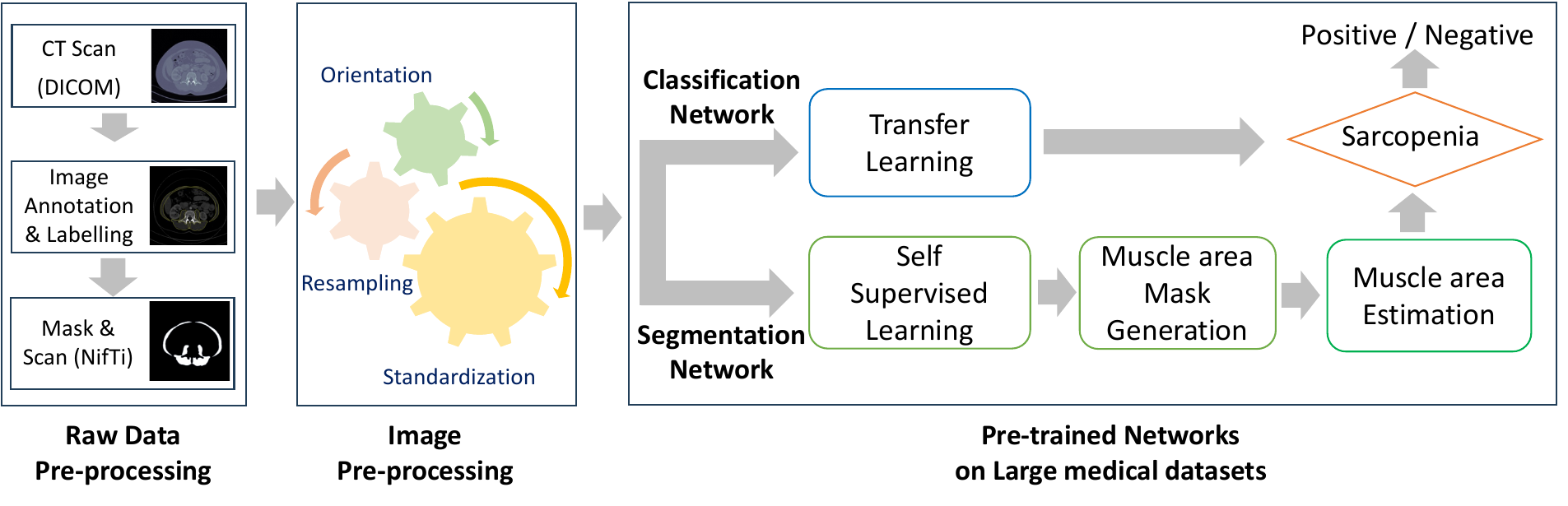}
        \caption{Data pre-processing and two deep learning approaches for automatically detecting sarcopenic and non-sarcopenic patients from CT scan images.}
        \label{fig:pipeline}
\end{figure}

We adopted two approaches for the automatic qualitative assessment of sarcopenia from CT scans at the level of L3. We used two end-to-end training pipelines. (i) an image classification approach using transfer learning for discriminating a CT image at the level of L3 as sarcopenic and non-sarcopenic. In this approach, a trained clinician has already carried out an SMA assessment and labelled data, and the model automatically classifies the CT scan into sarcopenic and non-sarcopenic states. (ii) A self-supervised learning approach for estimating the SMA from CT slices at the level of L3. This estimation of SMA leads to quantitative evaluation of a CT at L3 level as sarcopenic and non-sarcopenic. The overview of these two approaches is shown in Fig.~\ref{fig:pipeline}. 

\subsection{Freeman sarcopenia dataset and pre-processing methods}
{\color{black} Two types of datasets were prepared, one for the classification network and one for the self-supervised network. For the classification network, the dataset consists of two-dimensional bitmap images showing the section at L3. Each original image has a corresponding annotated image with outline markings of skeletal muscle, labelled at the Freeman Hospital by a trained clinician. For the self-supervised network, we collected 79 CT scans (mean age of 76 years, 52/79 male patients). 

These scans were anonymised, annotated, and labelled at the Freeman Hospital by clinicians with expertise in interpreting CT imaging. The scans were collected in the DICOM (Digital Imaging and Communications in Medicine) format, which is a series of files specific to a full scan of a single patient. These images together form a 3D image of the scan. The skeletal muscles scanned at L3 were annotated by manually outlining their area, an extremely time-consuming and cost-intensive task. Fig.~\ref{fig:annotatedscan} shows two examples of annotated CT scans at L3. Estimating the muscle area from the mask leads to assessing the sarcopenic and non-sarcopenic conditions.
\begin{figure}[h!]
        \centering
        \includegraphics[scale=0.6]{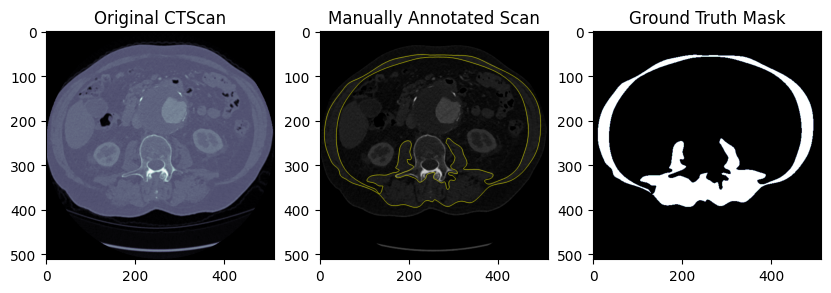} \\
        \includegraphics[scale=0.6]{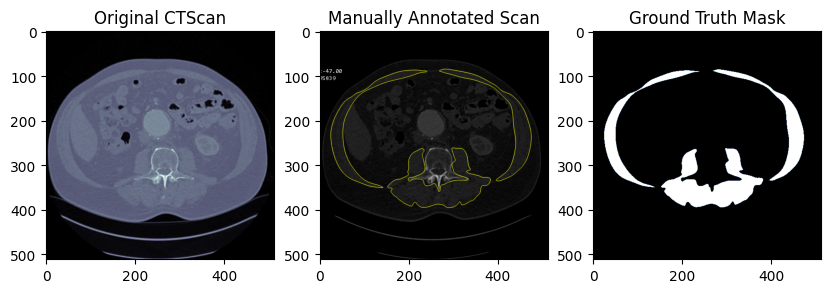}
        \caption{Two examples (two rows) of CT scans at the third lumbar vertebra (left column) along with annotated mask (centre column) and the mask produced from annotation (right column). The first row is a sarcopenic example, and the second row is a non-sarcopenic example.}
        \label{fig:annotatedscan}
\end{figure}
The corresponding annotated slice at the L3 vertebra level of the scan was then separated from these series. We converted the DICOM files into the NIfTI (Neuroimaging Informatics Technology Initiative) format to enable processing by the DL algorithm. 
}
{\color{black} 
The following steps were taken to pre-process the images: (i) \textit{Re-orientation} (RAS vs LAS). We process the data array axis of all scans to the same physical dimensions. That is, we orient the images/data array to align the first, second and third dimensions corresponding to coronal (from anterior to posterior), axial (from inferior to superior), and sagittal (from left to right). (ii) \textit{Standardisation of voxel spacing}. The physical coordinates of CT scans are mapped to voxel and are related by an affine matrix. This resizing or resampling was performed to standardize the voxel spacing. (iii) \textit{Normalisation of Hounsfield units} (HU). The intensity in CT scans is based on the HU; we clipped the values to retain between -175 to 250 (based on documentation of known SMA HU ranges~\cite{HUcutoff}) and then normalized between 0 and 1. (iv) \textit{Data augmentation}. Data augmentation was performed using rotation, flipping, padding and cropping.
}

\subsection{Image Classification Network}
\label{RadImageNet}
We use a transfer learning approach to image classification. Large pre-trained models that were trained on medical image datasets were fine-tuned using our target CT scan dataset of sarcopenic and non-sarcopenic patients. We use RadImageNet~\cite{mei2022radimagenet} trained models. RadImageNet is a large HIPAA-compliant medical image database containing 1.35 million annotated CT, MRI, and US images of musculoskeletal, neurologic, oncologic, gastrointestinal, endocrine, and pulmonary pathologic findings. The images contain ankle, foot, brain, hip, knee, shoulder, spine, pelvis, chest, thyroid and abdomen scans. We use a) Densenet121, b) InceptionResNetV2, and c) InceptionV3 models pre-trained on RadImageNet. We adopted a transfer learning approach of incremental fine-tuning of these models from the last classification layer (top layer) up to the input layer (unfreezing all layers). Our dataset has two classes: Negative (non-sarcopenic) and Positive (sarcopenic). The pre-trained models' last (classification) layer was modified with a two-node softMax.  Our dataset for the image classification network contained 79 labelled images (.bmp files), out of which an expert clinician labelled 18 images as Positive (or sarcopenic) and 61 images as Negative (or Non-sarcopenic). As the number of images are very limited, We created 10 datasets by shuffling images in to training data with 54 images (42 Negative and 12 Positive), Validation data with 12 images (9 Negative and 3 Positive), and test data with 13 images (10 Negative and 3 Positive. 
The optimum performance was achieved using the Adam optimizer with a learning rate of '0.00005' with batch size of 8 and utilizing early stopping. The images were pre-processed using corresponding network preprocess input function.
LIME and Gradcam are used for the explainability of these classification models.

\subsection{Self Supervised Network}
\label{subsec:SMITpara}
Despite we experiment with classification approach, the classification approach is not practical due to following challenges: (i)  it will have low number of labelled datasets, (ii) class will be always imbalanced as there will be less sarcopenic than non-sarcopenic CT scans, and (iii) qualitative assessemnet of CT scan do not inform about  sarcopenia or non-sarcopenia condition.  We used a self-supervised learning approach to mitigate these three challenges. the low number of labelled datasets. We use Self-distillation learning with the Masked Image modelling method to perform SSL for vision Transformers (SMIT) model~\cite{jiang2022self}. SMIT is a Vision Transformers (ViT) based model. ViT performs well despite image noise or contrast differences~\cite{dosovitskiy2021image}. Hence, these are suitable for use in medical segmentation. However, training in ViT requires a large number of labelled datasets, and this limitation can be overcome by self-supervised learning. We adopted and tuned SMIT models~\cite{jiang2022self} according to the requirements of our experiments and methodology. This network combines ViT with self-supervised learning using masked image modelling and self-distillation of teacher and student networks being trained concurrently. The student-teacher network is pre-trained for self-supervised learning using more than 3500 CT scans from the head, neck, chest, abdomen, lung, kidney, etc. After pre-training, only the student network is retained for further fine-tuning and testing. The student network was fine-tuned with our available limited CT scan data set. The student network was fine-tuned to produce masks and then the skeletal muscle area was calculated from the mask. 

Our dataset has 79 CT scans and corresponding annotated masks, with 18 scans as Positive (or sarcopenic) and 61 scans as Negative (or Non-sarcopenic). As the number of CT scans are very limited, We created 10 datasets by shuffling scans in to training data with 47  scans (of which 8 were sarcopenic and 39 were non-sarcopenic), 16 scans were used for validation (of which 6 were sarcopenic and 10 were non-sarcopenic), and 16 scans were used for testing (of which 3 were sarcopenic and 13 were non-sarcopenic). Once the muscle masks were predicted, the pixels represented by the masks were identified, which helped calculate skeletal muscle area utilizing the affine data of the CT scans. We used pre-trained student model weights and fine-tuned the models over 200 epochs using the AdamW optimizer with a learning rate of 0.0001 and a regularization weight of 0.00001. We saved the model after every 30 epochs so that we can observe the best performance during the testing.

\section{Results}
\label{sec:results}
We conducted experiments on Google Colab with A100 GPU and 40GB of GPU RAM to process the CT scan images and train the deep learning networks. The results of Image classification networks and self-supervised learning networks are as follows.

\subsection{Image classification network results} 
\begin{figure}[h!]
    \centering
        \includegraphics[scale=0.35]{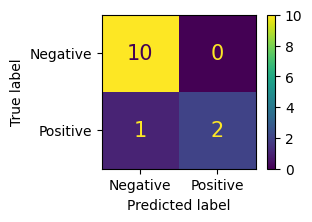}
        \includegraphics[scale=0.35]{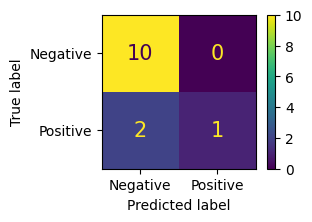}
        \includegraphics[scale=0.35]{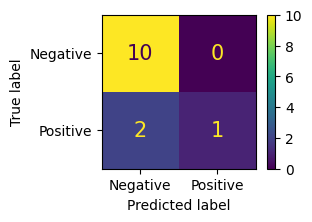}
        \includegraphics[scale=0.35]{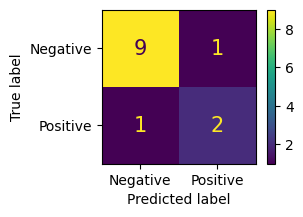}
        \\
        (a) Left to right: DenseNet, InceptionResNetV2, InceptionV3, and ResNet50
        
        \includegraphics[width=\textwidth]{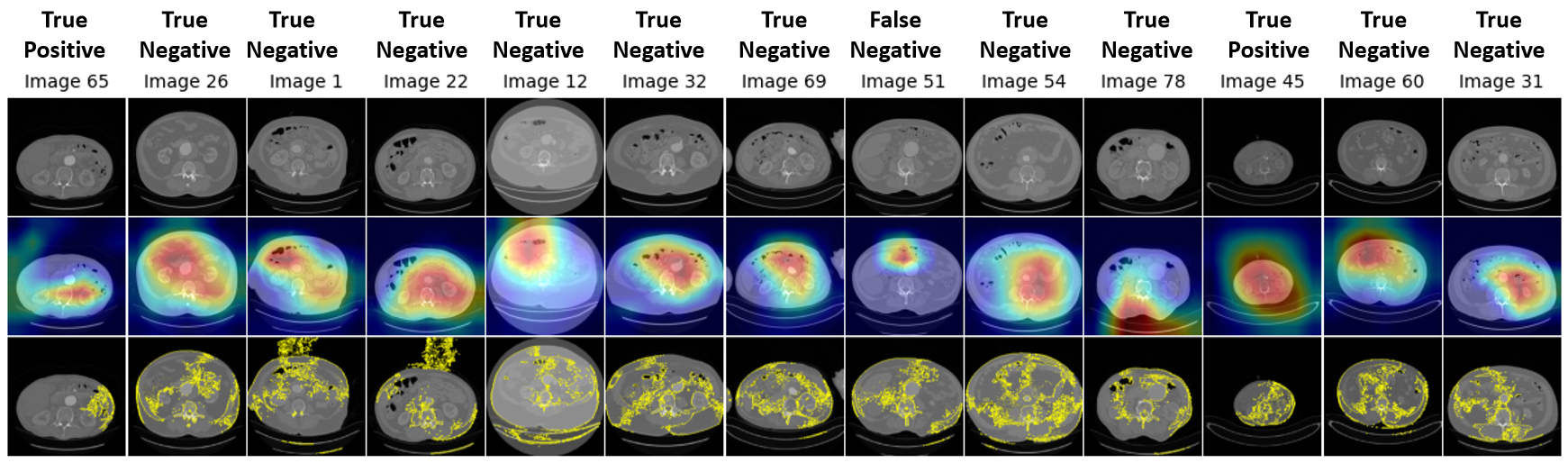}\\
        (b) DenseNet: The 8th image is predicted as a false negative.
        \includegraphics[width=\textwidth]{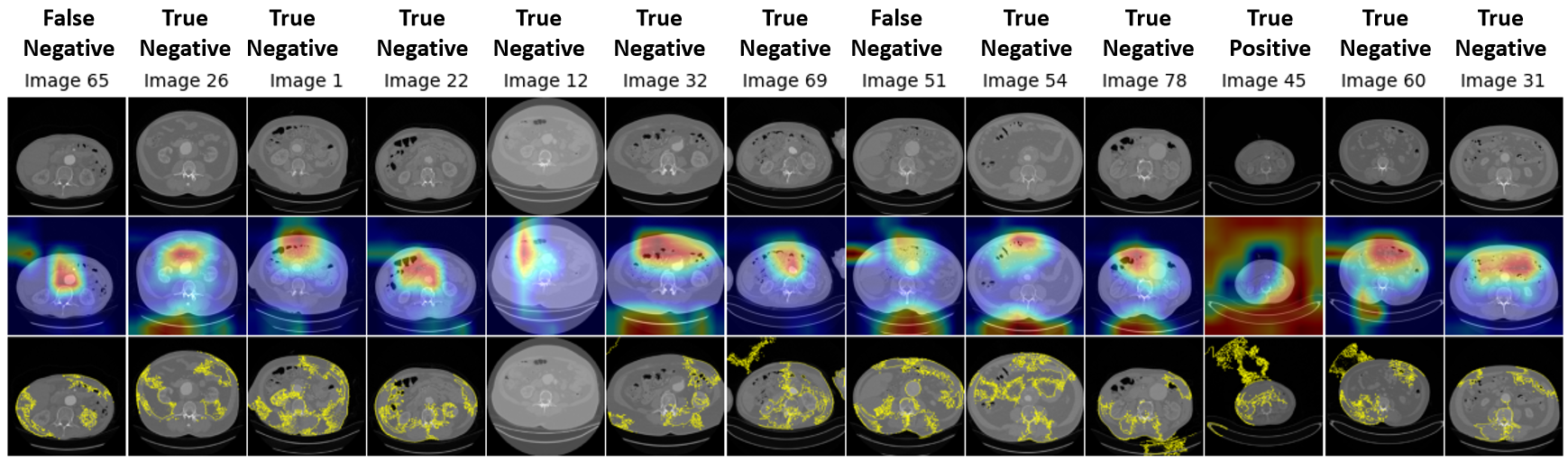}\\
        (c) InceptionResNetV2: The 1st and 8th images are predicted false negatives 
        
        \caption{Networks classification performance on test images. (a) Confusion metrics of the models on the test dataset. (b-c) LIME and GradCAM-based explanation, Original image (Top row), Gradcam map (Centre row), and LIME map (Bottom row).}
        \label{fig:rad_confusionmatrix_densenet}
\end{figure}

The pre-trained model was fine-tuned for each dataset of the 10 datasets in our experiments using InceptionV3, DenseNet121, ResNet50,  InceptionResNetV2 and each with layers configuration of unfreezeall, freezeall, and unfreezeTop10. The Table~\ref{tab:rad_auc_average_all} showcases the results based on which we selected the top four networks.
\begin{table}[h!]
 \setlength{\tabcolsep}{6pt}
  \centering
  \caption{Average and Standard Deviation of Accuracy across ten datasets}
  \begin{tabular}{lcc}
    \toprule
    %%\multicolumn{2}{c}{Category}                   \\
    %\cmidrule(r){1-2}
    Network             & Average AUC (10 bootstrap) & Std. Dev. of AUC \\
    \midrule
    InceptionV3\_unfreezeall     & 0.88    & 0.04    \\
    DenseNet121\_unfreezeall     & 0.88    & 0.07    \\
    ResNet50\_unfreezeall        & 0.83    & 0.04    \\
    IRV2\_unfreezeall            & 0.82    & 0.07    \\
    InceptionV3\_unfreezetop10   & 0.81    & 0.07    \\
    InceptionV3\_freezeall       & 0.81    & 0.06    \\
    IRV2\_freezeall              & 0.80    & 0.04    \\
    IRV2\_unfreezetop10          & 0.79    & 0.09    \\
    DenseNet121\_freezeall       & 0.79    & 0.03    \\
    DenseNet121\_unfreezetop10   & 0.78    & 0.08    \\
    ResNet50\_unfreezetop10      & 0.74    & 0.10    \\
    ResNet50\_freezeall          & 0.73    & 0.08    \\
    
    \bottomrule
  \end{tabular}
  
  \label{tab:rad_auc_average_all}
\end{table}
Table~\ref{tab:acctable} shows the test accuracy of the top four networks. The test accuracy of all of these networks was more than 80\%. However, out of the 13 test images, the models made critical errors on some test images (see Fig.~\ref{fig:rad_confusionmatrix_densenet}). IncentionNetV3 network, however, has classified all of the positive (sarcopenic) scans correctly but made two mistakes on non-sarcopenic (negative). %The test images where networks made errors (i.e., false positives and false negatives) are shown in Fig.~\ref{fig:IV3_fn}.
\begin{table}[h!]
 \setlength{\tabcolsep}{9pt}
 \caption{Test accuracy of three image classification networks and transfer learning approaches.}
  \centering
  \begin{tabular}{llrrr}
    \toprule
    %%\multicolumn{2}{c}{Category}                   \\
    \cmidrule(r){1-2}
    Pre-trained Network & Fine-tuning     & Accuracy & Precision & F1-Score  \\
    \midrule
    Densenet121         & Unfreeze all    & \textbf{0.92}    & \textbf{1}  & 0.80 \\
    InceptionResNetV2         & Unfreeze all    & 0.85    & 1  & 0.5  \\
    InceptionV3   & Unfreeze all    & 0.85    & 1    & 0.5  \\
    ResNet50            & Unfreeze all    & 0.85    & .67    & 0.67  \\
    \bottomrule
  \end{tabular}
  \label{tab:acctable}
\end{table}

%\begin{figure}[h!]
%    \centering
%        \includegraphics[scale=0.12]{media/Picture9b_IRV2_fp.png}~\quad
%        \includegraphics[scale=0.12]{media/Picture11c_DN_fp2.png}~\quad
%        \includegraphics[scale=0.12]{media/Picture17b_DN_fp3.png}     \\
%        (a) \\
%
%        \includegraphics[scale=0.12]{media/Picture9b_IRV2_fp.png}~\quad
%        \includegraphics[scale=0.12]{media/Picture10d_IV3_fp2.png}\\
% 	(b) 
%        
%       \includegraphics[scale=0.12]{media/Picture7d_IV3_fp1.png}~\quad
%        \includegraphics[scale=0.12]{media/Picture10d_IV3_fp2.png}\\
%        (c) \\
%        \caption{Networks classification performance on test images. (a) False negative: Positive (sarcopenic) examples are predicted as negative (non-sarcopenic) examples by DensNet121. (b) False negative (left): a positive (sarcopenic) example predicted as a negative (non-sarcopenic) example by InceptionResNetV2 and false positive (right): a negative (non-sarcopenic) example predicted as a positive example by InceptionResNetV2. (c) False positive: negative (non-sarcopenic) examples were predicted as positive by InceptionNetV3. InceptionNetV3 has zero false negatives.    
%        }
%        \label{fig:IV3_fn}
%\end{figure}

\subsection{Self-supervised network results}
\label{subsec:smitresult}
Unlike the image classification networks, where we ran into the risk of classifying into either of two classes, the SMA estimation approach offers a robust alternative. SMA estimation is an informative method for clinicians to infer sarcopenic and non-sarcopenic conditions. We fine-tuned the student network of a pre-trained SMIT network. The results of the SMA estimation are shown in Table \ref {tab:smitacc}. For further analysis of whether an estimated SMA is sarcopenic or non-sarcopenic, we compare the SMA to previously defined cutoffs~\cite{fumagalli2024automated,derstine2018skeletal}. As such, estimated SMA at L3 is classified as sarcopenic if the muscle area is below \qty{144}{\cm\squared} for a male or below \qty{92}{\cm\squared} for a female. Note that in Table~\ref{tab:smitacc}, we removed all descriptors of the personal identifier, even gender information, to make data fully anonymised. 
The figure~\ref{tab:smit_accuracy_10folds} showcases the dice accuracy across the ten folds, out of which we selected the best performing model to analyse further results.

\begin{table}[h!]
 \setlength{\tabcolsep}{2pt}
 
  \centering
  \caption{Average and Standard Deviation of Dice and area prediction error}
  \begin{tabular}{lcccccccccc}
    \toprule
    %%\multicolumn{2}{c}{Category}                   \\
    %\cmidrule(r){1-2}
    Statistics & Fold1 & Fold2 & Fold3 & Fold4 & Fold5 & Fold6 & Fold7 & Fold8 & Fold9 & Fold10 \\
    \midrule
    Average - Dice & 0.90 & 0.98 & 0.87 & 0.93 & 0.88 & 0.87 & 0.88 & 0.91 & 0.89 & 0.90   \\
    StDev - Dice & 0.04 & 0.05 & 0.05 & 0.02 & 0.06 & 0.05 & 0.06 & 0.04 & 0.05 & 0.05   \\
    Average- Area diff \% & 5.42 & 4.77 & -3.18 & 3.17 & -3.16 & -3.0 & 10.01 & 3.37 & 3.46 & 4.63   \\
    StDev- Area diff \% & 8.13 & 7.82 & 12.49 & 4.67 & 11.38 & 12.30 & 8.01 & 8.92 & 9.58 & 7.11   \\
    \bottomrule
  \end{tabular}
  
  \label{tab:smit_accuracy_10folds}
\end{table}

\begin{figure}[H]
  \centering
      \includegraphics[scale=0.3]{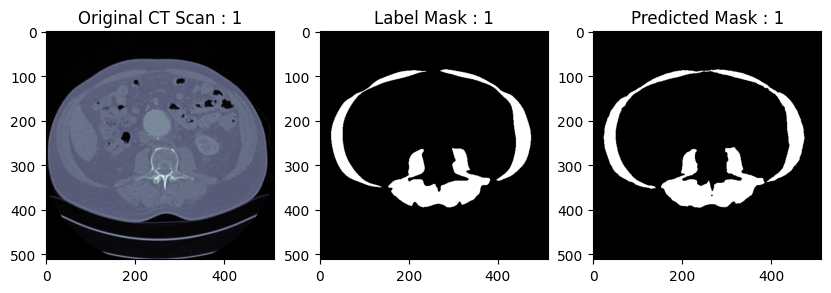} \\
      \includegraphics[scale=0.3]{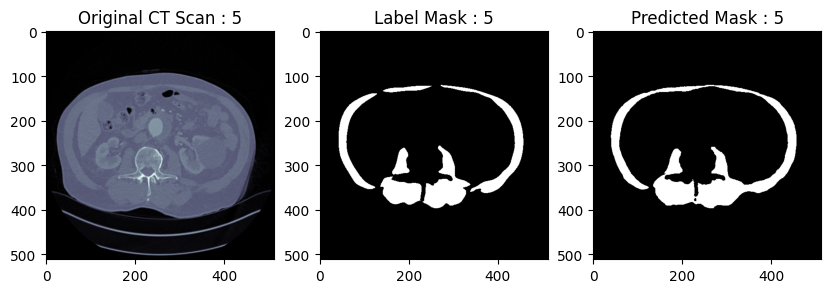} \\
      \includegraphics[scale=0.3]{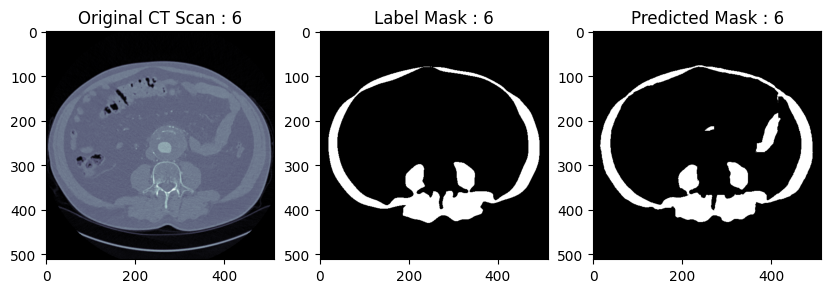} 
      \caption{CT scans of sarcopenic patients and corresponding predicted masks}
      \label{fig:SSL_postive_results}
\end{figure}

\begin{figure}[H]
  \centering
      \includegraphics[scale=0.3]{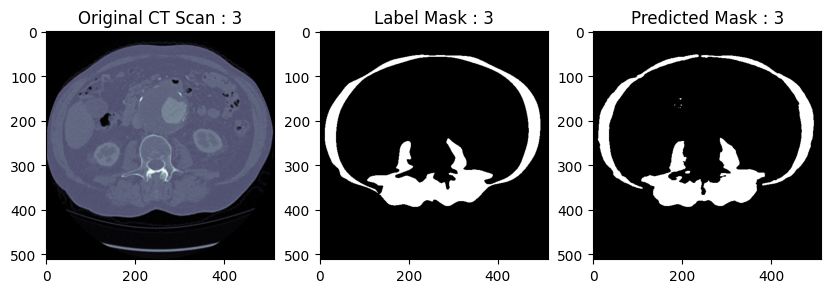} \\
      \includegraphics[scale=0.3]{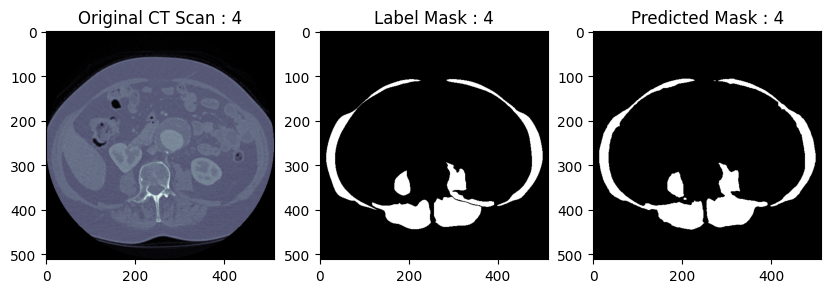} \\
      \includegraphics[scale=0.3]{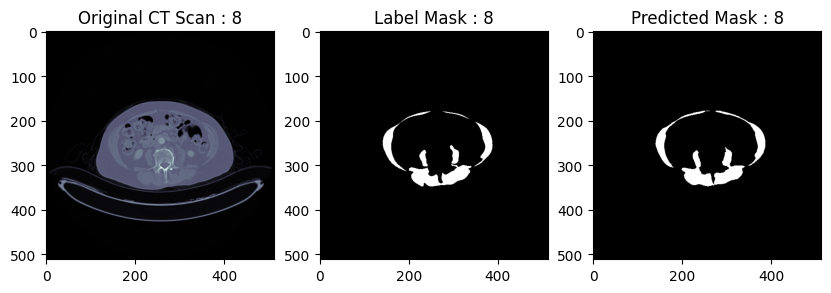} 
      \caption{CT scans of non-sarcopenic patients and corresponding predicted masks}
      \label{fig:SSL_negetive_results}
\end{figure}

The results of sarcopenia detection using a self-supervised network are significant, as the accuracy of prediction is 100\%, and the overall difference between actual and predicted muscle area is between 1\% to 7\% with a mean and median difference of 3\%. This makes this approach a potential tool for assisting clinicians. We also note that our work offers a higher Dice similarity coefficient (DSC) of 93\% than Gu et al.~\cite{gu2023detection} that achieves 90\%. However, Gu et al.~\cite{gu2023detection} did not release the dataset to which we can compare our results.  Fig.~\ref{fig:SSL_postive_results} and Fig.~\ref{fig:SSL_negetive_results} show the predicted masks of both positive and negative CT scans.
\begin{table}[h!]
\setlength{\tabcolsep}{4pt}
 \caption{Predicted SMA against SMA measurement by trained clinicians.}
  \centering
  \begin{tabular}{rcccrccc}
   \hline
    \multicolumn{2}{c}{Ground truth}  & & \multicolumn{3}{c}{Prediction} & & Error (\%) \\
    \cline{1-2}  \cline{4-6}
    SMA \unit{\cm\squared} & Sarcopenia & & Dice Score & SMA (\unit{\cm\squared}) & Sarcopenia & & Muscle area \\
    \hline
    180.1    & No & & 0.96 & 176.96   & No  & &  1.74\%   \\
    114.7     & No & & 0.89 & 115.3    & No  & &  0.53\%   \\
    163.42   & No & &  0.94 & 160.88   & No  & &  1.56\%   \\
    94.86    & No & &  0.92 & 199.38   & No  & &  2.32\%   \\
    158.31   & No & &  0.93 & 157.3    & No  & &  0.64\%   \\
    112.23   & No & &  0.92 & 108.46   & No  & &  3.36\%  \\
    106.82   & No & &  0.87 & 113.67   & No  & &  6.41\%   \\
    105.32    & No & &  0.91 & 99.92    & No  & &  5.13\%   \\
    149.61   & No & &  0.95 & 144.94   & No  & &  3.12\%   \\
    121.577  & No & &  0.93 & 118.90  & No  & &  2.19\%   \\
    97.18    & No & &  0.93 & 94.63    & No  & &  2.62\%   \\
    113.05   & No & &  0.95 & 104.98   & No  & &  7.14\%   \\
    156.92   & No & &  0.95 & 152.55   & No  & &  2.78\%   \\
    143.6    & Yes & &  0.93 & 135.5   & Yes  & &  5.64\%  \\
    117.33   & Yes & &  0.94 & 116.53  & Yes  & &  0.68\%   \\
    83.17    & Yes & &  0.92 & 79.09    & Yes  & &  4.91\%   \\
    \hline
  \end{tabular}
  \label{tab:smitacc}
\end{table}

\section{Conclusion}
\label{sec:con}
We address the challenging task of sarcopenia identification using computed tomography (CT) scans from a cohort of 79 patients collected and manually annotated by expert clinicians at the Freeman Hospital, Newcastle upon Tyne, UK. To support automated sarcopenia detection, we implemented both qualitative and quantitative approaches: (i) a qualitative approach using image classification to predict sarcopenia status directly and (ii) a quantitative method involving the estimation of skeletal muscle area (SMA) at the level of the third lumbar vertebra (L3). While the qualitative method could provide a straightforward binary answer with above 82\%  accuracy, including a model IncepatioNet with no false positive on test data, the quantitative assessment of SMA using a self-supervised learning approach offered 100\% accuracy with an average 3\% difference in estimating SMA between clinician assessment and algorithm assessment. The advantage of the self-supervised learning approach is two-fold: it requires less data for fine-tuning the model. It estimates SMA, alleviating the need for time-consuming and cost-intensive manual calculation of the area. Thus offering an effective assistive method for clinicians and radiologists. In future work, we will incorporate extensive studies for automatic identification of L3 level (currently done manually) and SMA measurement at different body parts, such as mid-thigh and mid-leg, for a comprehensive and fully automated Sarcopenia assessment using the same CT scans.

%\section*{Data availability}
%We will release the dataset to our university's open-source repository. However, due to the double-blind review policy, we are unable to do so at this stage. 

%%This project was considered to be a quality improvement project and therefore did not required formal ethical approval. The project was approved by Newcastle University.

%Bibliography
\bibliographystyle{splncs04}  
\bibliography{references}  

\end{document}